\documentclass[letter,journal]{IEEEtran}

\ifCLASSOPTIONcompsoc
  \usepackage[nocompress]{cite}
\else
  \usepackage{cite}
\fi
\ifCLASSINFOpdf
\else
\fi

\usepackage{amssymb}
\usepackage{graphicx}
\usepackage{booktabs}
\usepackage{hyperref}
\usepackage{amsmath}
\usepackage{algorithm,algpseudocode}
\usepackage{tikz}
\usepackage{pdfpages}
\usepackage{pgfplots}
\pgfplotsset{compat=newest}

\begin{document}
%
\title{Clustered FedStack: Intermediate Global Models with Bayesian Information Criterion}
%
%
%
%

\author{Thanveer Shaik,  Xiaohui Tao,  Lin Li, Niall Higgins, Raj Gururajan, Xujuan Zhou, Jianming Yong
\thanks{Thanveer Shaik, Xiaohui Tao are with 
the School of Mathematics, Physics \& Computing, University of Southern Queensland, Toowoomba, Queensland, Australia (e-mail: Thanveer.Shaik@usq.edu.au, Xiaohui.Tao@usq.edu.au).}
\thanks{Lin Li is with the School of Computer and Artificial Intelligence, Wuhan University of Technology, China (e-mail: cathylilin@whut.edu.cn)}
\thanks{Niall Higgins is with Royal Brisbane and Women’s Hospital and Queensland University of Technology, Australia (e-mail: Niall.Higgins@health.qld.gov.au)}

\thanks{Raj Gururajan is with School of Business, University of Southern Queensland, Australia  (e-mail: Raj.Gururajan@usq.edu.au).}
\thanks{Xujuan Zhou  is with School of Business, University of Southern Queensland, Australia  (e-mail: Xujuan.Zhou@usq.edu.au).}
\thanks{Jianming Yong  is with School of Business, University of Southern Queensland, Australia  (e-mail: Jianming.Yong@usq.edu.au).}
}

\IEEEtitleabstractindextext{%
\begin{abstract}
Federated Learning (FL) is currently one of the most popular technologies in the field of Artificial Intelligence (AI) due to its collaborative learning and ability to preserve client privacy. However, it faces challenges such as non-identically and non-independently distributed (non-IID) and data with imbalanced labels among local clients. To address these limitations, the research community has explored various approaches such as using local model parameters, federated generative adversarial learning, and federated representation learning. In our study, we propose a novel Clustered FedStack framework based on the previously published Stacked Federated Learning (FedStack) framework. The local clients send their model predictions and output layer weights to a server, which then builds a robust global model. This global model clusters the local clients based on their output layer weights using a clustering mechanism. We adopt three clustering mechanisms, namely K-Means, Agglomerative, and Gaussian Mixture Models, into the framework and evaluate their performance. We use Bayesian Information Criterion (BIC) with the maximum likelihood function to determine the number of clusters. The Clustered FedStack models outperform baseline models with clustering mechanisms. To estimate the convergence of our proposed framework, we use Cyclical learning rates.
\end{abstract}

}

\maketitle

\IEEEdisplaynontitleabstractindextext

%
\IEEEpeerreviewmaketitle

\ifCLASSOPTIONcompsoc
\IEEEraisesectionheading{\section{Introduction}\label{sec:introduction}}
\else
\section{Introduction}
\label{sec:introduction}
\fi

As AI techniques have matured, a vast amount of human data is being generated every second around the world. To manage this huge data, technology giant Google introduced a mechanism that trains a machine learning (ML) algorithm across multiple decentralized devices or servers without exchanging their local data samples. This is called Federated Learning (FL), which is also known as collaborative learning~\cite{bonawitz2021federated}. FL overcomes the issues of data privacy that exist in traditional centralized learning techniques where all device or server data is merged for analysis~\cite{shi2023federated}. FL has garnered significant attention since its introduction by Google as a ML technique for predicting users' input from Gboard (a keypad) on Android devices. This technique has been widely adopted in communication, engineering, and healthcare. However, medical institutes in particular possess a vast amount of patient data that may not be sufficient to train ML or deep learning models, and may even be biased due to a lack of data diversity. FL addresses this issue through its collaborative learning approach, where local models trained in each medical institute share their model weights with a global model stored in a shared server~\cite{peng2022evaluation}. This maintains data privacy, as the institute's data remains within its premises. The process can be used at the patient level to monitor their health status by predicting vital signs, such as heart rate and breathing, and classifying their physical activities. It enables personalized patient monitoring with enhanced data privacy.

A heterogeneous stacked FL, FedStack, was proposed by Shaik et al.~\cite{shaik2022fedstack} to overcome the problems of the traditional FL approach, while enabling personalized monitoring of patients' physical actions. The authors achieved state-of-the-art performances using different deep learning models as part of local and global clients. The FedStack approach is confined to building the global model by stacking local clients' predictions heterogeneously and allowing local clients to have different architectural models. However, it has a limitation of non-identically and independently distributed (non-IID) data, where the local clients' data distributions may be different. This can be addressed by allowing the global model to group the local clients based on their deep learning model output weights. To avoid any bias in grouping the local clients, unsupervised clustering methods can be adopted.

This study proposes a novel Clustered-FedStack framework to overcome FL's non-IID data challenge~\cite{arafeh2023data}. All models trained on local clients pass their predictions and output layer weights to the server, which builds a global server model based on the predictions received from the local models. Later, the global server model clusters local client models with output layer weights received and creates intermediate clustered models between local clients and the server. In this unsupervised process, the server model computes the cosine distance matrix among the local model output layer weights. To determine the number of clusters in this process, the BIC technique is adopted and maximum likelihood estimation is applied to the local model weights in the server. Three types of clustering techniques: centroid-based (k-Means), hierarchical (Agglomerative), and distribution-based (Gaussian Mixture Model) techniques are deployed. Cyclical learning rates are applied to estimate the convergence of the clustered models.

The proposed framework is evaluated with a human activity recognition (HAR) task using the publicly available sensor-based PPG-DALiA dataset~\cite{reiss2019deep}. The results show that clustered models have state-of-the-art performance in classifying human activities with the sensor data of 15 subjects. The performance of the clustered FedStack model is compared with four clustered FL baseline models, and the proposed model has outperformed the baseline models in all classification metrics. Moreover, the proposed framework can be scalable to Natural Language Processing (NLP) tasks. This has been evaluated on the drug review dataset~\cite{misc_drug_review_dataset_(drugs.com)_462}, where the intermediate clustered models performed better and could handle a huge number of local clients with non-IID data to achieve superconvergence. Thus, the proposed clustered FedStack framework can group local clients and overcome the non-IID challenge in FL. The contributions of the present study include the following:

\begin{itemize}
    \item A novel Clustered-FedStack framework is proposed to group local clients in an unsupervised approach and overcome the non-IID challenge in FL.
    \item Improved personalized modeling in FL by building intermediate clustered models between the global server model and local clients.
    \item Achievement of superconvergence of all clustered-FedStack models using Cyclical learning rates.
    \item A Clustered-FedStack approach that proves scalable for Natural Language Processing (NLP) tasks, effectively handling a high number of local clients with non-IID data.
\end{itemize}

Section~\ref{relatedworks} presents related works on FL and different aggregating techniques developed. Section~\ref{methods} presents the formulation of the research problem and the proposed Clustered-FedStack framework. In Section~\ref{Exp}, the proposed framework is evaluated in HAR and the results are discussed. The framework optimization with Cyclical learning rates is also presented in Section~\ref{Exp}. In Section~\ref{scalable}, we evaluate the scalability of the proposed framework using a NLP dataset. Section~\ref{conclude} concludes the paper.

\section{Related Works}\label{relatedworks}
Numerous studies have explored the aggregation of local model parameters in FL and passed them to the global model on the server. One of the first proposed aggregation techniques in FL is the Federated Averaging (FedAvg) algorithm, which uses the average function to aggregate local model weights and generate new weights to feed to the global model~\cite{mcmahan2017communication}. However, the FedAvg technique cannot optimize models if a client has a heterogeneous data distribution. To combat this, Arivazhagan et al.~\cite{arivazhagan2019federated} proposed FedPer, which has two layers: a base layer and a personalization layer. FedAvg trains the base layers, while the personalization layers are trained with stochastic gradient descent, helping to mitigate the ill effects of statistical heterogeneity. Wang et al.~\cite{wang2020federated} proposed Federated Matched Averaging (FedMA), which is a layer-wise approach that matches and merges nodes with the same weights, trains them independently, and communicates the layers to the global model.

Osmani et al.~\cite{osmani2022reduction} proposed a multi-level FL system for HAR, which includes a reconciliation step based on FL aggregation techniques such as FedAvg or Federated Normalized Averaging. Xiao et al.~\cite{xiao2021federated} proposed another FL system for HAR with enhanced feature extractions. They designed a Perceptive Extraction Network (PEN) with two networks: a featured network based on the convolutional block to extract local features, and a relation network based on Long Short-Term Memory (LSTM) and an attention mechanism to mine global relationships. Pang et al.~\cite{pang2022rule} proposed a rule-based collaborative framework (CloREF) that allows local clients to use heterogeneous local models. Tian et al.~\cite{9774841} discussed the limitations of traditional FL methods in heterogeneous IoT systems and proposed a novel Weight Similarity-based Client Clustering (WSCC) approach to address the non-IID challenge in FL. The WSCC approach involves splitting clients into different groups based on their data set distributions using an affinity-propagation-based method. Their proposed approach outperformed existing FL schemes under different non-IID settings, achieving up to 20\% improvements in accuracy without requiring extra data transmissions or additional models.

\begin{figure*}[ht]
    \centering
    \includegraphics[width=0.9\textwidth]{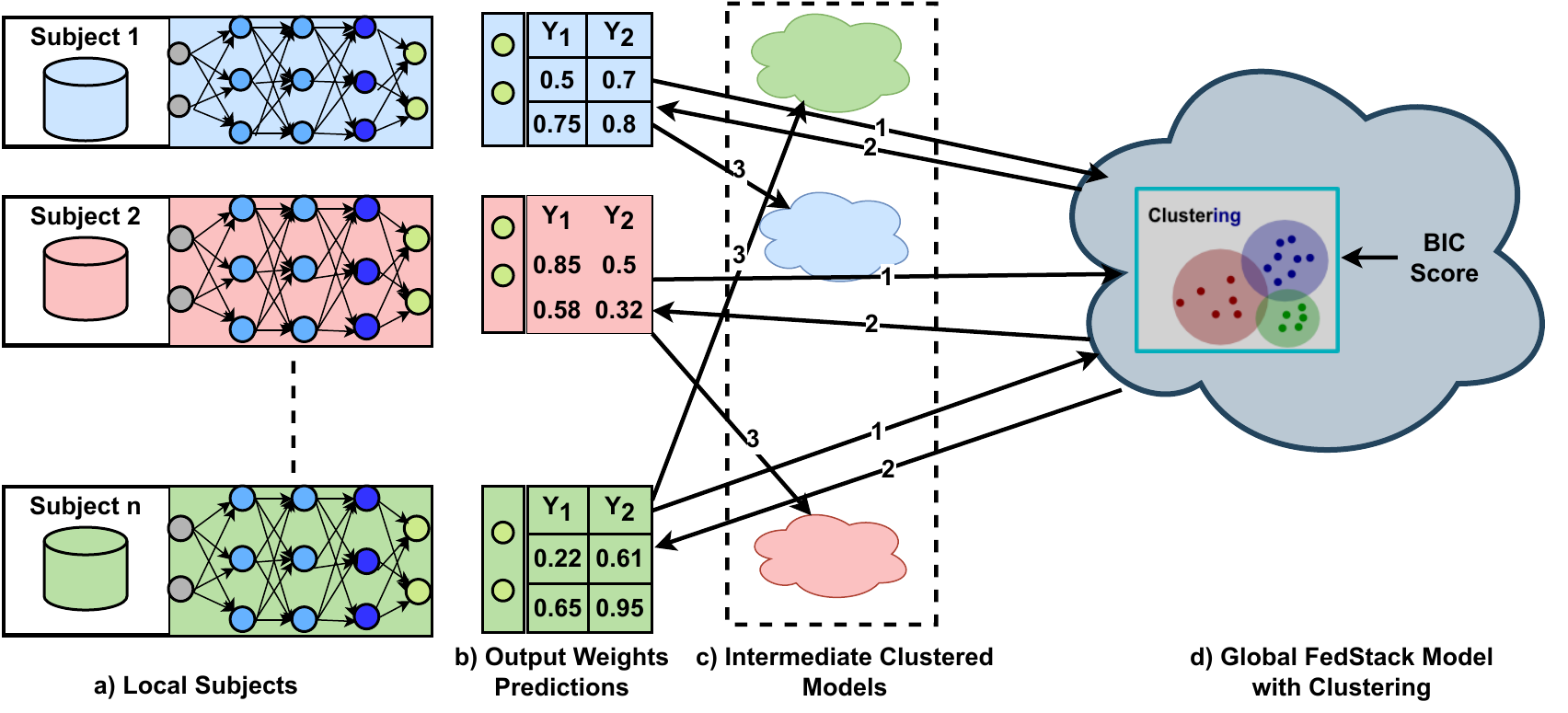}
    \caption{Clustered FedStack Model}
    \label{fig:clustered_fedstack}
\end{figure*}

\textbf{Federated Learning in HAR}
The increasing use of electronic assistive health applications such as smartwatches and activity trackers has led to the emergence of pervasive or ubiquitous computing, where devices can seamlessly exchange data with each other~\cite{alam2020fog}. Although this has the advantage of real-time tracking of human health changes, it is vulnerable to security breaches that compromise data privacy~\cite{dang2019survey}. The advancement of AI techniques as a whole is contributing to the massive amount of human data being generated worldwide every second. To handle such enormous data, Google introduced FL, which trains a ML algorithm across decentralized devices or servers without exchanging their local data samples. FL overcomes the data privacy issues associated with traditional centralized learning techniques, where all device and/or server data is merged for analysis~\cite{bonawitz2021federated}. Sannara et al.~\cite{ek2020evaluation} evaluated the performance of FL aggregation techniques like Federated Averaging (FedAvg), Federated Learning with Matched Averaging (FedMa), and Federated Personalization Layer (FedPer) against centralized training techniques. They used the CNN model to classify eight physical activities. Zhao et al.~\cite{zhao2020semi} designed an activity recognition system based on semi-supervised FL. Ouyang et al.~\cite{ouyang2021clusterfl} proposed the ClusterFL approach, which exploits the similarity of users' data to minimize the empirical loss of trained models. This improved Federated model accuracy and communication efficiency between local models and global models.

Local clients may have different data distributions, demographics, and model architectures. Passing all the local clients' parameters to build a robust global server model poses challenges such as label imbalance and non-IID. To identify hidden patterns or relationships among the local clients and overcome these challenges, unsupervised clustering techniques can be adopted to improve personalized learning in FL. This study proposes a clustered FL framework to overcome these identified challenges.

\section{Methodology}\label{methods}
To accommodate heterogeneous architectural models for local clients, we adopt the previously published FedStack framework by Shaik et al.~\cite{shaik2022fedstack}. This study extends the FedStack framework to the clustered-FedStack framework, facilitating the creation of heterogeneous multi-global FL models by clustering individual subjects with local models.

\subsection{Research Problem}
In this study, the research problem is to overcome the non-IID data challenge in a FL environment. Let $ S = \{s_{1}, s_{2}, \ldots, s_{N}\} $ be the set of subjects, where the data is non-IID. The objective is to divide subjects $ S $ into $ M $ clusters $ C = \{c_{1}, c_{2}, \ldots, c_{M}\} $, where each cluster $ c_{m} $ is a subset of subjects $ S $, $ c_{m} \subseteq S $. For each cluster $ c_{m} $, there exists a local model $ l_{m} $ that can be heterogeneous according to the subject's convenience. The predictions $ p_{m} $ from local models and their corresponding output layer weights are passed to a global model server $ g $. The training process for the global model $ g $ is shown in Equation~\ref{federated_equation}.

\begin{equation}\label{federated_equation}
    \text{train}(g) \leftarrow \sum_{m=1}^{M} c_{m} \leftarrow \sum_{m=1}^{M} l_{m}(p_{m})
\end{equation}

where: $\text{train}(g)$ refers to the training process for the Global Model $ g $ using local model predictions of cluster $ c_{m} $, and $ l_{m}(p_{m}) $ represents the local model $ l_{m} $ and its predictions $ p_{m} $ for each subject in the cluster $ c_{m} $.

\subsection{Clustered-FedStack Framework}
In the Clustered-FedStack framework, local clients train their models on private data and then forward their model predictions $p$ and output layer weights $Q$ to the global server model $g$ for training, as shown in Fig.~\ref{fig:clustered_fedstack}. The figure's arrow numbers indicate the framework execution order. After receiving the local model predictions and output layer weights, the global server model determines the number of clusters using the BIC score. It then clusters the local clients based on their output layer weights. For each label $i$ in local model training, an output neuron without a successor is configured to gather the computed and accumulated values from the local model's input and hidden layers. The output neuron value $q_{i}$ is calculated using Equation~\ref{output_neuron}, with inputs $x_{i}$, weights $w_{i}$, and bias $b$ for a local model $l_{n}$. By computing all output neuron values, the local model $l_{n}$ predictions $p$ can be estimated using Equation~\ref{predictions}.

\begin{equation}\label{output_neuron}
q_{i} = l_{n}(b, x_{i}, w_{i})
\end{equation}
\begin{equation}\label{predictions}
p  = l_{n}(b + \sum_{i=1}^{n} x_{i} \cdot w_{i})
\end{equation}

Output neuron values for each local model $l_{n}$ are consolidated into a single set $Q$ using Equation~\ref{singleset}. This procedure is repeated for all local models based on their output layer values, forming a large set $\mathcal{Q}$ as defined in Equation~\ref{largeset}.

\begin{equation}\label{singleset}
Q = \{q_{1}, q_{2}, q_{3}, \dots, q_{n}\}
\end{equation}
\begin{equation}\label{largeset}
\mathcal{Q} = \{Q_{l_{1}}, Q_{l_{2}}, \dots, Q_{l_{n}}\}
\end{equation}

\subsubsection{Clustering Technique}
Given the set $\mathcal{Q}$ from Equation~\ref{largeset}, where each element of the set represents the values of a local model's $l_{n}$ output layer, the goal is to divide $\mathcal{Q}$ into $k$ clusters, where $k \leq n$, represented by $C = \{c_{1}, c_{2}, \dots, c_{k}\}$. There are various techniques that can be applied to clustering, including centroid-based, hierarchical, and distribution-based methods. The general objective of these methods is to minimize the within-cluster sum of squared differences or a related measure of dissimilarity, as described in Equation~\ref{cluster}. The notation $\underset{C}{\text{arg min}}$ refers to finding the set of clusters $C$ that minimizes the following expression, where the ``arg min'' stands for the argument of the minimum, i.e., the specific value of the variable that results in the lowest possible value of the given function.

\begin{equation}\label{cluster}
\underset{C}{\text{arg min}} \left( \sum_{i=1}^{k} \sum_{x \in C_{i}} {||x - c_{i}||}^2 \right)
\end{equation}

Here, $C$ represents the set of clusters, $x$ is a data point, and $c_{i}$ is the representative point, such as a centroid. The term ${||\cdot||}$ represents a distance measure.

Cosine similarity is utilized to assign each local model's output neuron set to a specific cluster, considering the angle between output neuron sets of two local model $l_{n}$ as $Q_{l_{1}}$ and $Q_{l_{2}}$, the cosine similarity can be estimated using the Equation~\ref{cosine}.

\begin{equation}\label{cosine}
S_{C}(Q_{l_{1}}, Q_{l_{2}}) = \frac{Q_{l_{1}} \cdot Q_{l_{2}}}{ ||Q_{l_{1}}||~||Q_{l_{2}||}}
\end{equation}


\subsubsection{Bayesian Information Criterion}
The proposed Clustered-FedStack technique enables the global server model to access local models' predictions and layer weights. However, using an unsupervised method to determine the number of clusters in local models is challenging. The BIC technique is utilized to overcome this. BIC calculates its value based on a clustering model $\mathcal{M}$'s maximum likelihood function $M_{L}$, representing the probability that the layer weights data fits the clustering model~\cite{xiang2008design}. This is shown in Equation~\ref{ml}. BIC values balance the maximum likelihood estimation against the number of model parameters $m_{p}$, seeking a model with the fewest parameters that can accurately explain the data clusters, as in Equation~\ref{bic}.

\begin{equation}\label{ml}
M_{L}(\mathcal{M}) = -2 \ln(\mathcal{L}) + m_{p} \ln(n)
\end{equation}
\begin{equation}\label{bic}
BIC = -2 \ln(\mathcal{L}) + m_{p} \ln(n) = M_{L}(\mathcal{M})
\end{equation}

The BIC values for each clustering model are compared with the minimum BIC value indicating the optimal clustering model. This process ensures that the global model converges by configuring a suitable number of clusters for local models, resulting in a consolidated global model that represents heterogeneous subject models.

\subsection{Clustered-FedStack Algorithm}

\begin{algorithm}[!ht]
\caption{Proposed Clustered-FedStack Algorithm}\label{alg:cap}
\scriptsize
\begin{algorithmic}[1]
\Require
    \Statex Subjects set $S = \{s_{1},s_{2},\dots,s_{n}\}$
    \Statex Local AI models $M = \{m_{1},m_{2},\dots,m_{m}\}$
    \Statex Labels set $K = \{1,2,\dots,k\}$
    \Statex Global Server Model $g$
\Ensure Classification probabilities of labels $K$ for each intermediate cluster model $C$
\State Initialization: 
    \Statex $D$: Dataset for training
    \Statex $D'$: Unseen Dataset for testing
    \Statex $W = \emptyset$: Set to collect weights
    \Statex $CM = \emptyset$: Set for clustered models
\State $stack = \left\{{\{m^{K}_i, m^{w}_i\}},{\{m^{K}_j, m^{w}_j\}},{\{m^{K}_k, m^{w}_k\}}\right\}$; \Comment{Predictions and weights of local AI models}
\For {$m \in M$}
    \State $g^{train} \leftarrow stack$;
    \State $g^{test} \leftarrow D'$;
\EndFor
\For {$m\in G(M)$}
    \State Collect weights of $m$: $W \leftarrow \{m, w\}$;
\EndFor
\State \textbf{Determine Clusters:} 
\State Compute BIC scores of $CM \geq M$;
\State $CM \leftarrow \min(\text{BIC})$;
\State Compute cosine distance among $\{\{m_{1},w_{1}\},\{m_{2},w_{2}\},\dots,\{m_{m},w_{m}\}\}$;
\State \textbf{Assignment:}
\For {c in $C$} 
        \State $c \leftarrow \underset{C}{\text{arg min}} \left( \sum_{i=1}^{k} \sum_{x \in C_{i}} {||x - c_{i}||}^2 \right)$; 
        \State $CM \leftarrow c$;
\EndFor
\State \textbf{Return} $CM$;
\end{algorithmic}
\end{algorithm}

Algorithm~\ref{alg:cap} presents the proposed Clustered-FedStack process in detail. Line 1 initializes empty sets to collect output layer weights and clustered models, and datasets $D$ and $D'$ for evaluating the global server model. Lines 2-7 detail the FedStack process, where local client model predictions and weights are passed to the global server model $g$ for training and testing. Lines 8-10 present the iteration through all local model weights in $g$ to collect their output layer weights. Lines 11-12 detail the determination of the number of clusters to be formed from the weights $W$ set. Line 13 computes the cosine distance among all the local model weights collected. Lines 14-19 explain the clustering process for all the local models, based on Lines 11-13.

\begin{figure}[!ht]
    \centering
    \includegraphics[width=0.9\columnwidth]{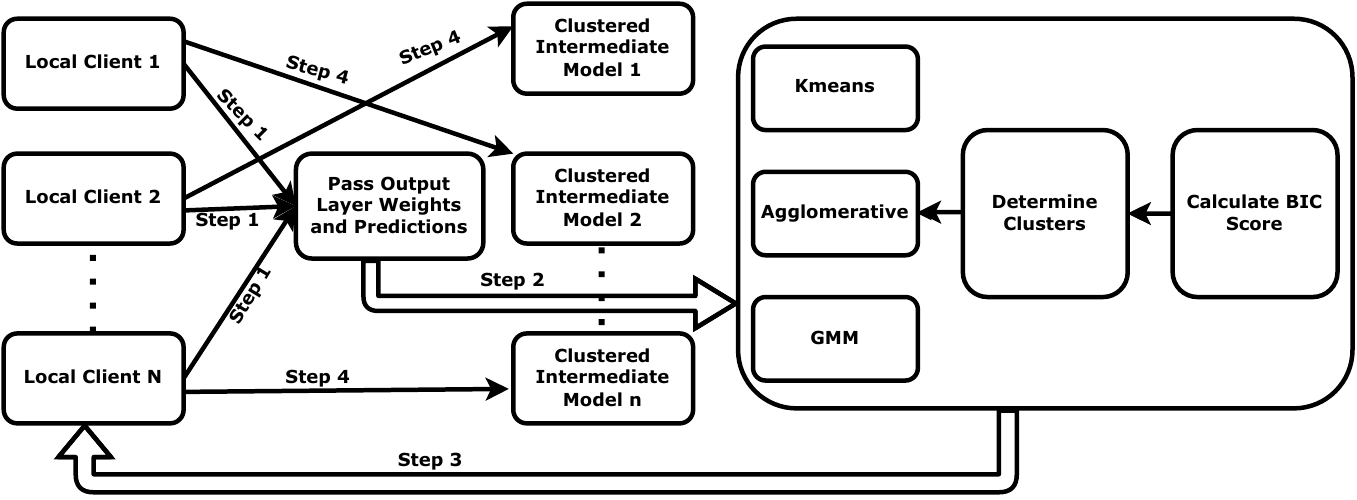}
    \caption{Experimental Design of the proposed framework}
    \label{fig:expdesign}
\end{figure}

\section{Experiments on Clustered FedStack in HAR}\label{Exp}
Conventional FL methods assume that the data distribution is consistent among all clients~\cite{yang2022personalized}. However, this assumption may not be valid in FL, as data heterogeneity can be present~\cite{shang2022fedic}. This limitation forces clients to have identical data distribution and architectural models to build global models. FedStack~\cite{shaik2022fedstack} addressed the issue of identical architectural models in FL. The goal of this study is to extend the FedStack framework by introducing intermediate clustered models to address the non-IID challenge in FL.

In this study, the objective is to overcome the non-IID challenge in FL. To achieve this, the proposed Clustered-FedStack algorithm is applied to the domain of human activity recognition, where patients' physical activity is classified. The non-IID data distribution of the dataset used in the experiment is presented. The proposed methodology involves passing the output layer weights and predictions of local clients to the global model, which then calculates unsupervised clustering of the local model layer weights to group the local clients and establish clustered intermediate models. The experimental design is presented in Figure~\ref{fig:expdesign}. The evaluation results compare the performance of the proposed framework to the baseline models and show clustering results leading to clustered FedStack models. Furthermore, the convergence of the clustered FedStack models is analyzed using Cyclical learning rates.

\subsection{Dataset}
The proposed Clustered-FedStack algorithm was evaluated on the HAR problem, which involves classifying patients' physical activity. The PPG-DALiA~\cite{reiss2019deep} dataset, which is publicly accessible and cited in \cite{reiss2019deep}, was utilized for this study. This dataset includes physiological and motion data gathered from 15 participants as they engaged in a diverse array of activities, closely mirroring real-life conditions. The data was collected from both a wrist-worn (Empatica E4) and a chest-worn (RespiBAN) device, and includes 11 attributes such as 3-Dimensional (3D) acceleration data, Electrocardiogram (ECG), respiration, Blood Volume Pulse (BVP), Electrothermal Activity (EDA), and body temperature. The 3D acceleration data was labelled with eight different physical activities.

\begin{table}[ht]
\centering
\caption{Non-IID Data}
\label{tab:non_iid_table}
\resizebox{\columnwidth}{!}{%
\begin{tabular}{ll*{8}{c}}
\toprule
\textbf{Local Clients} & \textbf{Distribution} & \textbf{1} & \textbf{2} & \textbf{3} & \textbf{4} & \textbf{5} & \textbf{6} & \textbf{7} & \textbf{8} \\
\midrule
Subject 1 & 27724 & 2800 & 1148 & 1380 & 1648 & 3556 & 9420 & 3016 & 4756 \\ \midrule
Subject 2 & 22712 & 2400 & 1068 & 1216 & 1548 & 3680 & 4880 & 2756 & 5164 \\ \midrule
Subject 3 & 26900 & 2400 & 1740 & 1172 & 1516 & 3640 & 8640 & 2952 & 4840 \\ \midrule
Subject 4 & 26528 & 2280 & 2092 & 1312 & 1900 & 4028 & 7580 & 2376 & 4960 \\ \midrule
Subject 5 & 26924 & 2400 & 1860 & 1160 & 1728 & 3320 & 9020 & 2356 & 5080 \\ \midrule
Subject 6 & 11812 & 2532 & 1720 & 1236 & 2132 & 4192 & 9020 & 0 & 0 \\ \midrule
Subject 7 & 28580 & 2472 & 1624 & 1096 & 2012 & 4140 & 9700 & 2836 & 4700 \\ \midrule
Subject 8 & 23992 & 2400 & 1648 & 1292 & 1680 & 3080 & 7200 & 1924 & 4768 \\ \midrule
Subject 9 & 26212 & 2400 & 1932 & 1140 & 2216 & 3820 & 7368 & 2356 & 4980 \\ \midrule
Subject 10 & 28424 & 2392 & 1868 & 1220 & 1952 & 3748 & 8336 & 4328 & 4580 \\ \midrule
Subject 11 & 28052 & 2400 & 1828 & 1296 & 1960 & 3440 & 9632 & 2616 & 4880 \\ \midrule
Subject 12 & 23680 & 2408 & 1936 & 1120 & 1920 & 3560 & 5840 & 2116 & 4780 \\ \midrule
Subject 13 & 26996 & 2420 & 1988 & 1160 & 1992 & 3588 & 8112 & 2836 & 4900 \\ \midrule
Subject 14 & 25584 & 2432 & 1824 & 1300 & 2008 & 3816 & 6924 & 2460 & 4820 \\ \midrule
Subject 15 & 23504 & 2444 & 1676 & 1416 & 1620 & 3140 & 5760 & 2636 & 4812 \\ \bottomrule

\end{tabular}}

\end{table}

\subsection{Non-IID Data}
Tab.~\ref{tab:non_iid_table} shows the distribution and activity of local clients in a FL scenario with non-IID data. Each row represents a client, and each column represents a feature. The ``Distribution'' column shows the number of data points available at each client, which varies across clients, indicating non-IID in the dataset. The remaining columns represent different activities that are each related to the type of data collected or the task being performed. For instance, ``Activity 1'' to ``Activity 8'' could be different types of sensor readings or behavioral data collected from different sources. The non-IID nature of this data could potentially impact the performance of the FL algorithm since the data distribution across clients is not uniform, and the model may not generalize well to all clients. Therefore, special attention must be given to handling the non-IID data in FL, by using the technique of personalized FL to improve model performance for each client's unique data distribution.

\subsection{Data Modeling}
In data modeling, three AI models were chosen: Artificial Neural Networks (ANN), Convolutional Neural Networks (CNN), and Bidirectional Long Short-Term Memory (BiLSTM) models, due to their state-of-the-art performances in FL works~\cite{xiao2021federated} and activity classification~\cite{wang2019deep}. Each subject trained with one of the chosen models locally and passed their predictions and local model output layer weights to the global server model. The proposed framework clustered the global model without any private information about local clients, based on the output layer weights.

\begin{figure}[ht]
    \centering
    \includegraphics[width=0.9\columnwidth]{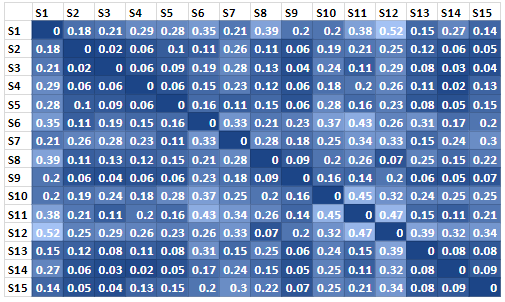}
    \caption{Cosine distance among local clients}
    \label{fig:cosine}
\end{figure}

\subsection{Baseline Models}
\begin{itemize}
\item \textbf{ClusterFL~\cite{ouyang2021clusterfl}:} A clustering-based FL system for the HAR application. The ClusterFL approach captures the intrinsic clustering relation among local clients and minimizes the training loss.
\item \textbf{FL+HC~\cite{briggs2020federated}:} A hierarchical clustered FL system to separate clusters of clients based on the similarity of their local updates to the global server model.
\item \textbf{HypCluster~\cite{mansour2020three}:} A hypothesis-based clustering with a stochastic Expectation-Maximization (EM) algorithm adopted for the FL approach, where local clients partition into a certain number of clusters and then the model finds the best hypothesis for each cluster.
\item \textbf{Dynamic Clustering~\cite{kim2021dynamic}:} A three-phased data clustering algorithm, namely, generative adversarial network-based clustering, cluster calibration, and cluster division, designed to overcome the fixed shape of clusters, data privacy breaches, and non-adaptive numbers of clusters.
\end{itemize}

\begin{figure}[ht]

\begin{tikzpicture}
\begin{axis}[ width=\columnwidth,
            height=5cm,   xlabel={BIC Score},    ylabel={No of Clusters},    tick label style={font=\footnotesize},    label style={font=\small},    title style={font=\small},    xmin=-500, xmax=0,    ymin=0, ymax=10,    xtick={-500,-400,-300,-200,-100,0},    ytick={1,2,3,4,5,6,7,8,9},    legend style={font=\small},    legend pos=north west,    ]
    
\addplot[color=blue,mark=square*] coordinates {
(-298.4662333639485,1)
(-423.6594571834727,2)
(-468.5431601193207,3)
(-342.2485203791746,4)
(-416.1944019368898,5)
(-363.91075805030107,6)
(-273.42855409799563,7)
(-201.97443100059866,8)
(-113.33712291374741,9)
};

\end{axis}
\end{tikzpicture}
    \caption{BIC score to determine the number of clusters}
    \label{fig:bic_score}
\end{figure}
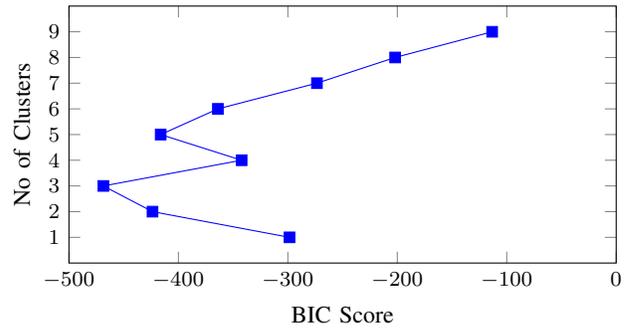

\subsection{Results Analysis}

\subsubsection{Clustering Results}
Before clustering, the cosine distance among all 15 local models trained on clients is calculated to check their similarity in terms of the models' output layer weights, as shown in Fig.~\ref{fig:cosine}. The matrix heatmap ranges on a scale from 0 to 0.6 where 0 shows no cosine distance between the client output layer values, and 0.6 shows the maximum cosine distance.

\begin{figure}[!ht]
    \centering
    \includegraphics[width=0.85\columnwidth]{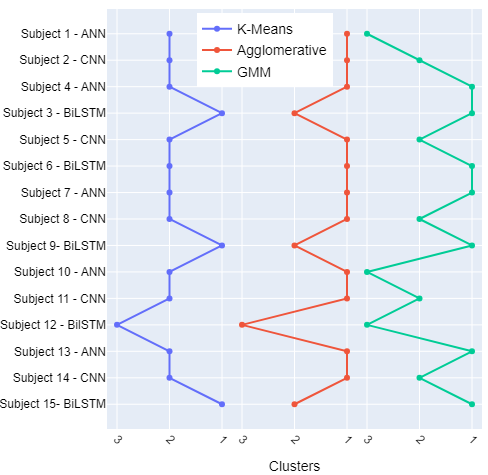}
    \caption{Clustering Results}
    \label{fig:cluster_results}
\end{figure}

The proposed Clustered-FedStack algorithm employed the BIC approach to calculate the maximum likelihood function on the output layer weights received from the local client models by the global server model, as shown in Fig.~\ref{fig:clustered_fedstack}. This process determines the number of clusters among the 15 local clients. Fig.~\ref{fig:bic_score} shows that the lowest BIC score corresponds to three clusters in the global server model. After determining the clusters, three clustering techniques were applied: centroid-based clustering (K-Means)~\cite{jain2010data}, hierarchical clustering (Agglomerative)~\cite{sellami2020fused}, and distribution-based clustering (Gaussian Mixture Model (GMM))~\cite{lucke2019k}. Fig.~\ref{fig:cluster_results} shows that K-Means and Agglomerative clustering produced similar groups of local client models, while GMM clustering grouped all CNN models into the second cluster and distributed other ANN and BiLSTM models in the first and third clusters.

\subsubsection{Clustered FedStack Model Performances}
After determining the clusters, each cluster of local clients passes their output layer weights to an intermediate Clustered-FedStack model, situated between the local clients and the global server model, as shown in Fig.~\ref{fig:clustered_fedstack}. This approach reduces the load on the global server model and groups similar local models for more efficient AI results. The three clustering techniques generate three Clustered-FedStack models each, and their performance in HAR is shown in Tab.~\ref{tab:clustered_fedstack}. All nine Clustered-FedStack intermediate global models generated from the clustering techniques have performed well in the HAR task. K-Means and agglomerative clustering, having similar clustering results, showed similar classification accuracy in HAR. While comparing the results, the GMM Clustered FedStack models, which are distribution-based, exhibited slightly better accuracy than the other two clustered models.

\begin{table}[]
\large
\caption{Clustered FedStack model accuracy in HAR}
\label{tab:clustered_fedstack}
\resizebox{\columnwidth}{!}{
\begin{tabular}{@{}c|ccc|ccc|ccc@{}}
\toprule
 & \multicolumn{3}{c|}{\textbf{\begin{tabular}[c]{@{}c@{}}K-Means\\ Clusters\end{tabular}}} & \multicolumn{3}{c|}{\textbf{\begin{tabular}[c]{@{}c@{}}Agglomerative \\ Clusters\end{tabular}}} & \multicolumn{3}{c}{\textbf{\begin{tabular}[c]{@{}c@{}}GMM \\ Clusters\end{tabular}}} \\ \midrule
\textbf{Activity} & \multicolumn{1}{c}{\textbf{Cluster 1}} & \multicolumn{1}{c}{\textbf{Cluster 2}} & \textbf{Cluster 3} & \multicolumn{1}{c}{\textbf{Cluster 1}} & \multicolumn{1}{c}{\textbf{Cluster 2}} & \textbf{Cluster 3} & \multicolumn{1}{c}{\textbf{Cluster 1}} & \multicolumn{1}{c}{\textbf{Cluster 2}} & \textbf{Cluster 3} \\ \midrule
\textbf{Sitting} & \multicolumn{1}{c}{0.99} & \multicolumn{1}{c}{0.96} & 0.95 & \multicolumn{1}{c}{0.97} & \multicolumn{1}{c}{0.99} & 0.95 & \multicolumn{1}{c}{0.99} & \multicolumn{1}{c}{0.99} & 0.99 \\ \midrule
\textbf{\begin{tabular}[c]{@{}c@{}}Ascending and\\ descending stairs\end{tabular}} & \multicolumn{1}{c}{0.92} & \multicolumn{1}{c}{0.96} & 0.94 & \multicolumn{1}{c}{0.96} & \multicolumn{1}{c}{0.92} & 0.94 & \multicolumn{1}{c}{0.92} & \multicolumn{1}{c}{0.92} & 0.92 \\ \midrule
\textbf{Table soccer} & \multicolumn{1}{c}{0.96} & \multicolumn{1}{c}{0.95} & 0.95 & \multicolumn{1}{c}{0.97} & \multicolumn{1}{c}{0.95} & 0.95 & \multicolumn{1}{c}{0.95} & \multicolumn{1}{c}{0.96} & 0.97 \\ \midrule
\textbf{Cycling} & \multicolumn{1}{c}{0.94} & \multicolumn{1}{c}{0.97} & 0.98 & \multicolumn{1}{c}{0.95} & \multicolumn{1}{c}{0.96} & 0.98 & \multicolumn{1}{c}{0.96} & \multicolumn{1}{c}{0.93} & 0.96 \\ \midrule
\textbf{Driving a car} & \multicolumn{1}{c}{0.89} & \multicolumn{1}{c}{0.93} & 0.99 & \multicolumn{1}{c}{0.89} & \multicolumn{1}{c}{0.95} & 0.99 & \multicolumn{1}{c}{0.95} & \multicolumn{1}{c}{0.93} & 0.97 \\ \midrule
\textbf{Lunch break} & \multicolumn{1}{c}{0.87} & \multicolumn{1}{c}{0.86} & 0.92 & \multicolumn{1}{c}{0.87} & \multicolumn{1}{c}{0.89} & 0.92 & \multicolumn{1}{c}{0.9} & \multicolumn{1}{c}{0.9} & 0.91 \\ \midrule
\textbf{Walking} & \multicolumn{1}{c}{0.91} & \multicolumn{1}{c}{0.90} & 0.89 & \multicolumn{1}{c}{0.90} & \multicolumn{1}{c}{0.92} & 0.89 & \multicolumn{1}{c}{0.91} & \multicolumn{1}{c}{0.91} & 0.92 \\ \midrule
\textbf{Working} & \multicolumn{1}{c}{0.92} & \multicolumn{1}{c}{0.96} & 0.95 & \multicolumn{1}{c}{0.97} & \multicolumn{1}{c}{0.97} & 0.95 & \multicolumn{1}{c}{0.96} & \multicolumn{1}{c}{0.92} & 0.97 \\ \bottomrule
\end{tabular}}
\end{table}

\subsubsection{Baseline Models Comparison}
The proposed framework was compared against four other baseline models in FL approaches with clustering. All models were trained using 3D acceleration data for HAR tasks, and their evaluation results are presented in Tab.~\ref{tab:baseline_comp}. As K-Means and hierarchical clustering techniques generate similar clusters from the 15 local client models, the table shows three clustered models (Clustered FedStack 1, Clustered FedStack 2, Clustered FedStack 3) built based on K-Means and hierarchical clustering, and three clustered models (Clustered FedStack 7, Clustered FedStack 8, Clustered FedStack 9) built based on the GMM model. The Table presents the mean of four metrics: balanced accuracy, precision, recall, and F1-score in classifying eight activities for six intermediate clustered models. The proposed approach outperformed all other baseline models in terms of all the metrics.

\begin{table}[]
\centering
\caption{Baseline Models Comparison}
\label{tab:baseline_comp}
\scriptsize
\resizebox{0.8\columnwidth}{!}{%
\begin{tabular}{@{}ccccc@{}}
\toprule
\textbf{Model}                & \textbf{Balanced Accuracy} & \textbf{Precision} & \textbf{Recall} & \textbf{F1-Score} \\ \midrule
ClusterFL~\cite{ouyang2021clusterfl}                     & 0.93              & 0.78               & 0.86            & 0.82              \\ \midrule
FL+HC~\cite{briggs2020federated}                       & 0.94              & 0.85               & 0.89            & 0.83              \\ \midrule
HypCluster~\cite{mansour2020three}                    & 0.9               & 0.65               & 0.56            & 0.65              \\ \midrule
Dynamic Clustering~\cite{kim2021dynamic}            & 0.92              & 0.86               & 0.75            & 0.76              \\ \midrule
\textbf{Clustered FedStack 1} & \textbf{0.98}     & \textbf{0.95}      & \textbf{0.91}   & \textbf{0.93}     \\ \midrule
\textbf{Clustered FedStack 2} & \textbf{0.96}     & \textbf{0.89}      & \textbf{0.9}    & \textbf{0.89}     \\ \midrule
\textbf{Clustered FedStack 3} & \textbf{0.94}     & \textbf{0.91}      & \textbf{0.92}   & \textbf{0.91}     \\ \midrule
\textbf{Clustered FedStack 7} & \textbf{0.95}     & \textbf{0.92}      & \textbf{0.91}   & \textbf{0.91}     \\ \midrule
\textbf{Clustered FedStack 8} & \textbf{0.98}     & \textbf{0.94}      & \textbf{0.93}   & \textbf{0.93}     \\ \midrule
\textbf{Clustered FedStack 9} & \textbf{0.97}     & \textbf{0.96}      & \textbf{0.95}   & \textbf{0.95}     \\ \bottomrule
\end{tabular}}
\end{table}
\subsection{Convergence Analysis}\label{opti}
The optimization of the proposed Clustered FedStack framework is estimated using Cyclical learning rates~\cite{smith2017cyclical} for convergence. The performance of the intermediate Clustered FedStack models shown in Fig.~\ref{fig:clustered_fedstack} is optimized using the Learning Rate ($\alpha$) of the deep learning models. In the Cyclical learning rates process, the $\alpha$ values are cycled with an initial learning rate of $0.00001$ and a maximum learning rate of $0.001$, and stochastic gradient descent is performed. A scale function is defined to control the change from the initial learning rate to the maximal learning rate and back to the initial learning rate. The scale function, a lambda function shown in Equation\ref{scale_function}, scales the initial amplitude by half with each cycle.

\begin{equation}\label{scale_function}
    lambda~~{x}:\frac{1}{(2^{(x-1)})}
\end{equation}

\begin{figure}[!ht]
    \centering
    \includegraphics[width=\columnwidth]{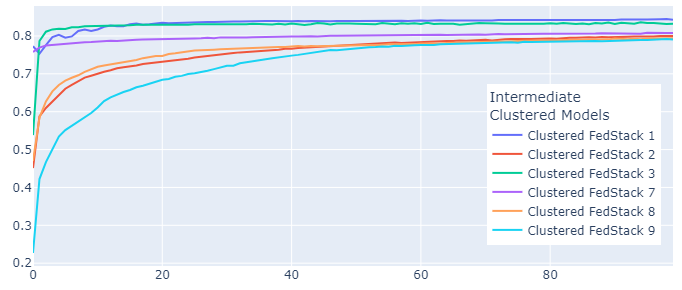}
    \caption{Convergence of intermediate Clustered FedStack models on PPG-DALiA under the Cyclical Learning Rates}
    \label{fig:convergence_plot}
\end{figure}

Fig.~\ref{fig:convergence_plot} presents the convergence curves of six intermediate Clustered FedStack models from the three clustering techniques proposed in this study. The intermediate clustered models built based on K-Means and Agglomerative clustering converge faster than the clustered models built based on GMM clustering. There is not much difference in the number of epochs required for each clustered model to converge. All six models converge in less than 50 epochs. The results show that the proposed Clustered FedStack framework can be implemented with centroid-based, hierarchical or distribution-based clustering. The Clustered FedStack models built based on any of these clustering techniques converge quickly in 50 epochs.

\section{Experiments on Clustered FedStack Scalability in NLP Tasks}\label{scalable}
The scalability of the proposed Clustered FedStack model was rigorously assessed through a targeted evaluation. For this purpose, the drug review dataset~\cite{misc_drug_review_dataset_(drugs.com)_462}, containing reviews and ratings, was utilized. This comprehensive dataset encompasses 3677 distinct drugs and 916 different medical conditions. The aim of this experiment is to classify drug ratings (1-10) based on input data such as medical conditions. In alignment with the clustering methodologies proposed in the Clustered FedStack framework, the GMM clustering was employed to perform the clustering of 2191 drugs, resulting in 78 unique clusters. The cosine distance of 200 local clients (drugs) is presented in the Supplementary Material. 

The performance comparisons of different clustering models, including the top 10 variations of the Clustered FedStack model, are presented in Tab.~\ref{tab:model_performance_drug}. The metrics evaluated include accuracy, precision, recall, and F1-score for classifying drug ratings. Four baseline models are included: ClusterFL, FL+HC, HypCluster, and Dynamic Clustering. Their performances are relatively consistent, with accuracy ranging from 0.89 to 0.93. The Clustered FedStack models demonstrated superior performance, with notable improvements in all evaluated metrics. The accuracy for these variations ranged from 0.94 to a perfect 1, highlighting the efficiency and robustness of the model. The first five Clustered FedStack models exhibited particularly impressive results, achieving almost perfect or perfect accuracy. The precision, recall, and F1-score also showcased strong consistency and harmony, reflecting the model's ability to balance both false positives and false negatives.

\begin{table}[]
\centering
\scriptsize
\caption{Clustered FedStack Performance in Classification of Drug Ratings}
\label{tab:model_performance_drug}
\resizebox{0.8\columnwidth}{!}{%
\begin{tabular}{@{}crrrr@{}}
\toprule
\textbf{Model} & \multicolumn{1}{c}{\textbf{Accuracy}} & \multicolumn{1}{c}{\textbf{Precision}} & \multicolumn{1}{c}{\textbf{Recall}} & \multicolumn{1}{c}{\textbf{F1-Score}} \\ \midrule
\multicolumn{1}{c}{ClusterFL} & \multicolumn{1}{r}{0.92} & \multicolumn{1}{r}{0.8} & \multicolumn{1}{r}{0.88} & \multicolumn{1}{r}{0.82} \\ \midrule
\multicolumn{1}{c}{FL+HC} & \multicolumn{1}{r}{0.93} & \multicolumn{1}{r}{0.87} & \multicolumn{1}{r}{0.91} & \multicolumn{1}{r}{0.83} \\ \midrule
\multicolumn{1}{c}{HypCluster} & \multicolumn{1}{r}{0.89} & \multicolumn{1}{r}{0.66} & \multicolumn{1}{r}{0.57} & \multicolumn{1}{r}{0.65} \\ \midrule
\multicolumn{1}{c}{Dynamic Clustering} & \multicolumn{1}{r}{0.91} & \multicolumn{1}{r}{0.88} & \multicolumn{1}{r}{0.77} & \multicolumn{1}{r}{0.76} \\ \midrule
\multicolumn{1}{c}{\textbf{Clustered FedStack 1}} & \multicolumn{1}{r}{0.99} & \multicolumn{1}{r}{0.92} & \multicolumn{1}{r}{0.93} & \multicolumn{1}{r}{0.91} \\ \midrule
\multicolumn{1}{c}{\textbf{Clustered FedStack 2}} & \multicolumn{1}{r}{0.98} & \multicolumn{1}{r}{0.91} & \multicolumn{1}{r}{0.92} & \multicolumn{1}{r}{0.91} \\ \midrule
\multicolumn{1}{c}{\textbf{Clustered FedStack 3}} & \multicolumn{1}{r}{1} & \multicolumn{1}{r}{0.96} & \multicolumn{1}{r}{0.95} & \multicolumn{1}{r}{0.95} \\ \midrule
\multicolumn{1}{c}{\textbf{Clustered FedStack 4}} & \multicolumn{1}{r}{0.97} & \multicolumn{1}{r}{0.94} & \multicolumn{1}{r}{0.93} & \multicolumn{1}{r}{0.93} \\ \midrule
\multicolumn{1}{c}{\textbf{Clustered FedStack 5}} & \multicolumn{1}{r}{0.98} & \multicolumn{1}{r}{0.97} & \multicolumn{1}{r}{0.94} & \multicolumn{1}{r}{0.96} \\ \midrule
\multicolumn{1}{c}{\textbf{Clustered FedStack 6}} & \multicolumn{1}{r}{1} & \multicolumn{1}{r}{0.97} & \multicolumn{1}{r}{0.93} & \multicolumn{1}{r}{0.95} \\ \midrule
\multicolumn{1}{c}{\textbf{Clustered FedStack 7}} & \multicolumn{1}{r}{0.99} & \multicolumn{1}{r}{0.98} & \multicolumn{1}{r}{0.97} & \multicolumn{1}{r}{0.97} \\ \midrule
\multicolumn{1}{c}{\textbf{Clustered FedStack 8}} & \multicolumn{1}{r}{0.98} & \multicolumn{1}{r}{0.97} & \multicolumn{1}{r}{0.94} & \multicolumn{1}{r}{0.96} \\ \midrule
\multicolumn{1}{c}{\textbf{Clustered FedStack 9}} & \multicolumn{1}{r}{0.96} & \multicolumn{1}{r}{0.93} & \multicolumn{1}{r}{0.94} & \multicolumn{1}{r}{0.93} \\ \midrule
\multicolumn{1}{c}{\textbf{Clustered FedStack 10}} & \multicolumn{1}{r}{0.94} & \multicolumn{1}{r}{0.93} & \multicolumn{1}{r}{0.92} & \multicolumn{1}{r}{0.91} \\\bottomrule
\end{tabular}}
\end{table}

These results underscore the scalability and effectiveness of the Clustered FedStack model across local clients with non-IID data. The model's scalability and adaptability are evident, maintaining high levels of accuracy and F1-scores regardless of the local clients' variation. This highlights the Clustered FedStack model's potential in managing large and intricate datasets like drug reviews and ratings, validating both its resilience and relevance to real-world applications.

\begin{figure}[!ht]
\centering
\includegraphics[width=\columnwidth]{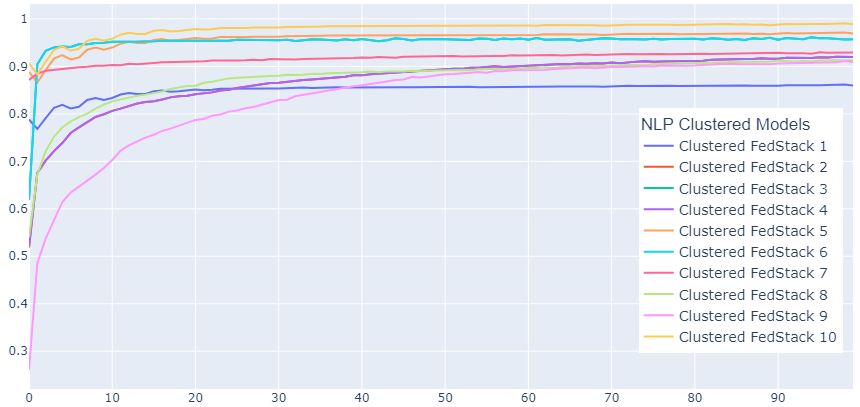}
\caption{Convergence of Intermediate Clustered FedStack Models on Drug Review Dataset under Cyclical Learning Rates}
\label{fig:cov_scalability}
\end{figure}

The convergence of the proposed Clustered FedStack on the drug review dataset has been assessed, as shown in Fig.~\ref{fig:cov_scalability}. The line chart presents a convergence pattern that denotes accuracy in the y-axis across 100 epochs in the x-axis. The values for Clustered FedStack 1 exhibited consistent growth, starting at 0.7882 and reaching 0.8556 by epoch 40. Similarly, other clustered FedStacks demonstrated a progressive increase in values across epochs, such as Clustered FedStack 2, which advanced from 0.5188 to 0.8811, signifying a gradual strengthening of the model. These convergence trends shed light on the efficiency and efficacy of the iterative learning process. Variations in convergence rates among different stacks were observed, reflecting the distinct characteristics of each clustered FedStack. These findings suggest a general trend of convergence towards higher values, though occasional oscillations and fluctuations were detected in specific iterations. This in-depth analysis offers valuable insights into the behavior of clustered FL systems, potentially opening new avenues for enhanced optimization strategies and a more profound understanding of convergence mechanisms within distributed ML frameworks.

\section{Conclusion}\label{conclude}
In the present study, a novel framework named Clustered-FedStack was introduced, designed to cluster local clients within the FL paradigm based on the weights of their output layers. This methodology was devised to address the non-IID challenge inherent to FL. It is important to acknowledge certain limitations of the proposed framework, notably its incompatibility with the application on local clients utilizing conventional Machine Learning models for the training of private data. Moreover, the global server model's process of clustering local clients operates on an unsupervised basis, without access to specific information about local clients, depending solely on the local model rather than client demographics. In light of these considerations, future investigations should aim to develop strategies for the dynamic clustering of local clients, taking into account meta-information that pertains to client similarities.

\bibliographystyle{ieeetr}
\bibliography{ieee}

\vspace{40cm}  

\includepdf[pages=-]{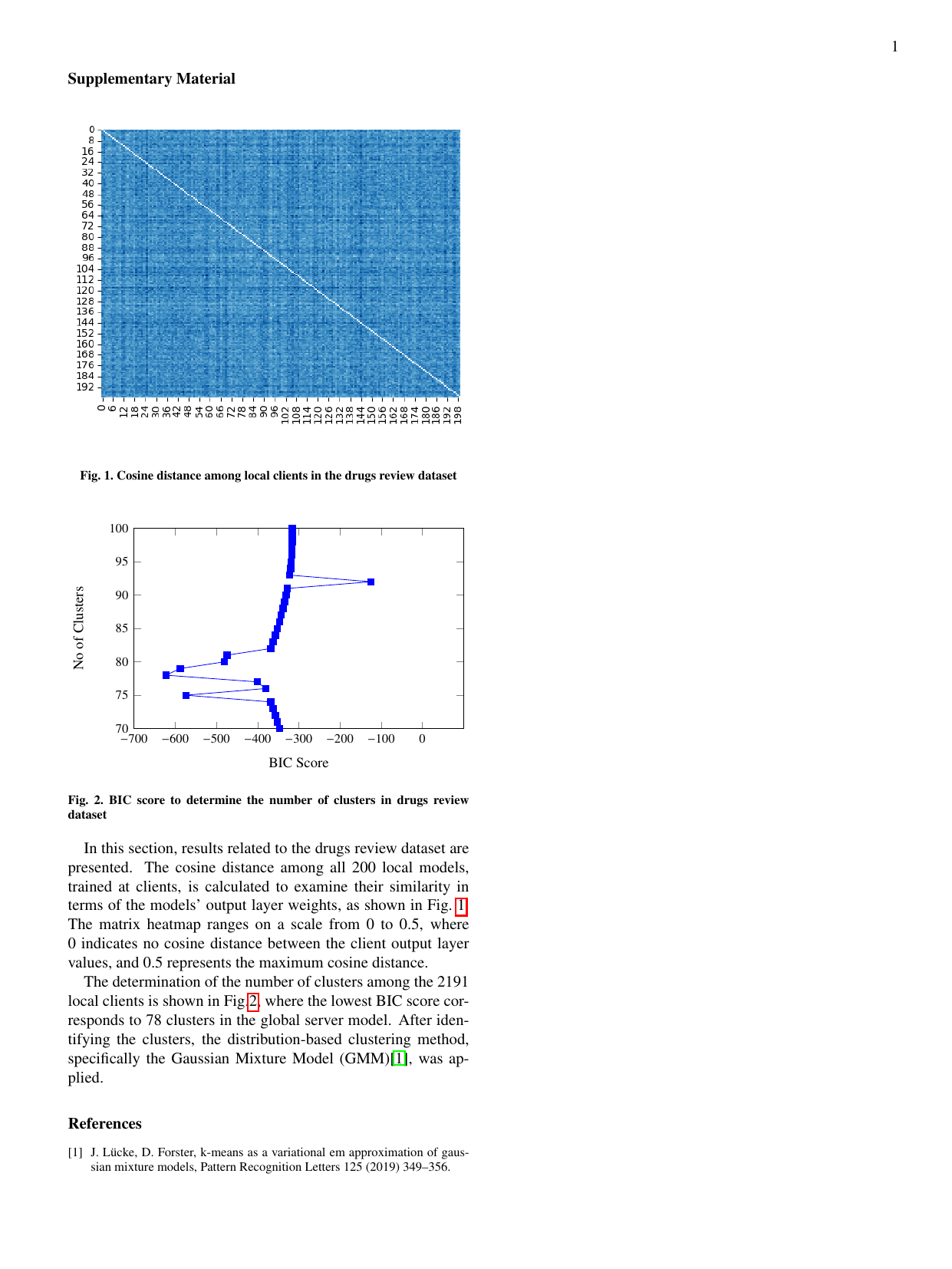}

\end{document}